\documentclass[11pt,a4paper]{article}
\usepackage[hyperref]{acl2020}
\usepackage{times}
\usepackage{latexsym}

\usepackage[utf8]{inputenc} 
\usepackage[T1]{fontenc}    
\usepackage{microtype}      
\usepackage{lipsum}
\usepackage{listings}

\usepackage{multirow}
\usepackage{xspace}
\usepackage{makecell}
\usepackage{colortbl}

\usepackage{graphicx}
\graphicspath{ {images/} }

\usepackage{color}
\definecolor{darkgreen}{rgb}{0,0.5,0}

\newcommand\model{\textsc{Quark}\xspace} 

\newcommand{\hpqa}{HotpotQA\xspace}
\newcommand{\roberta}{RoBERTa\xspace}
\newcommand{\ourlm}{\emph{BERT-Large-Cased}\xspace}

\newcommand{\namecite}[1]{\citeauthor{#1}~\shortcite{#1}}

\newcommand\T{\rule{0pt}{2.6ex}}       
\newcommand{\selectmod}{\rowcolor{purple!15}}

\def\aclpaperid{x}
\aclfinalcopy
\title{A Simple Yet Strong Pipeline for {HotpotQA}}

\author{
  Dirk Groeneveld$^\dag$ \and Tushar Khot$^\dag$ \and Mausam$^\ddag$ \and Ashish Sabharwal$^\dag$\\
  \ \\
  $^\dag$ Allen Institute for AI, Seattle, WA, U.S.A.\\
  \texttt{\small dirkg,tushark,ashishs@allenai.org} \\
  $^\ddag$ Indian Institute of Technology, Delhi, India\\
  \texttt{\small mausam@cse.iitd.ac.in}
}

\date{}

\begin{document}
\maketitle

\begin{abstract}
State-of-the-art models for multi-hop question answering typically augment large-scale language models like BERT with additional, intuitively useful capabilities such as named entity recognition, graph-based reasoning, and question decomposition. However, does their strong performance on popular multi-hop datasets really justify this added design complexity? Our results suggest that the answer may be no, because even our simple pipeline based on BERT, named \model, performs surprisingly well. Specifically, on \hpqa, \model outperforms these models on both question answering and support identification (and achieves performance very close to a \roberta model). Our pipeline has three steps: 1) use BERT to identify potentially relevant sentences \emph{independently} of each other; 2) feed the set of selected sentences as context into a standard BERT span prediction model to choose an answer; and 3) use the sentence selection model, now with the chosen answer, to produce supporting sentences. The strong performance of \model resurfaces the importance of carefully exploring simple model designs before using popular benchmarks to justify the value of complex techniques.
\end{abstract}

\section{Introduction}

Textual Multi-hop Question Answering (QA) is the task of answering questions by combining information from multiple sentences or documents. This is a challenging reasoning task that requires QA systems to identify relevant pieces of information in the given text and learn to compose them to answer a question. To enable progress in this area, many datasets~\cite{wikihop,Talmor2018TheWA,hotpotqa,qasc} and models~\cite{decomprc,dfgn,sae} with varying complexities have been proposed over the past few years. Our work focuses on \hpqa~\cite{hotpotqa}, which contains 105,257 multi-hop questions derived from two Wikipedia paragraphs, where the correct answer is a span in these paragraphs or yes/no. 

\begin{figure}
    \centering
    \includegraphics[width=0.75\columnwidth]{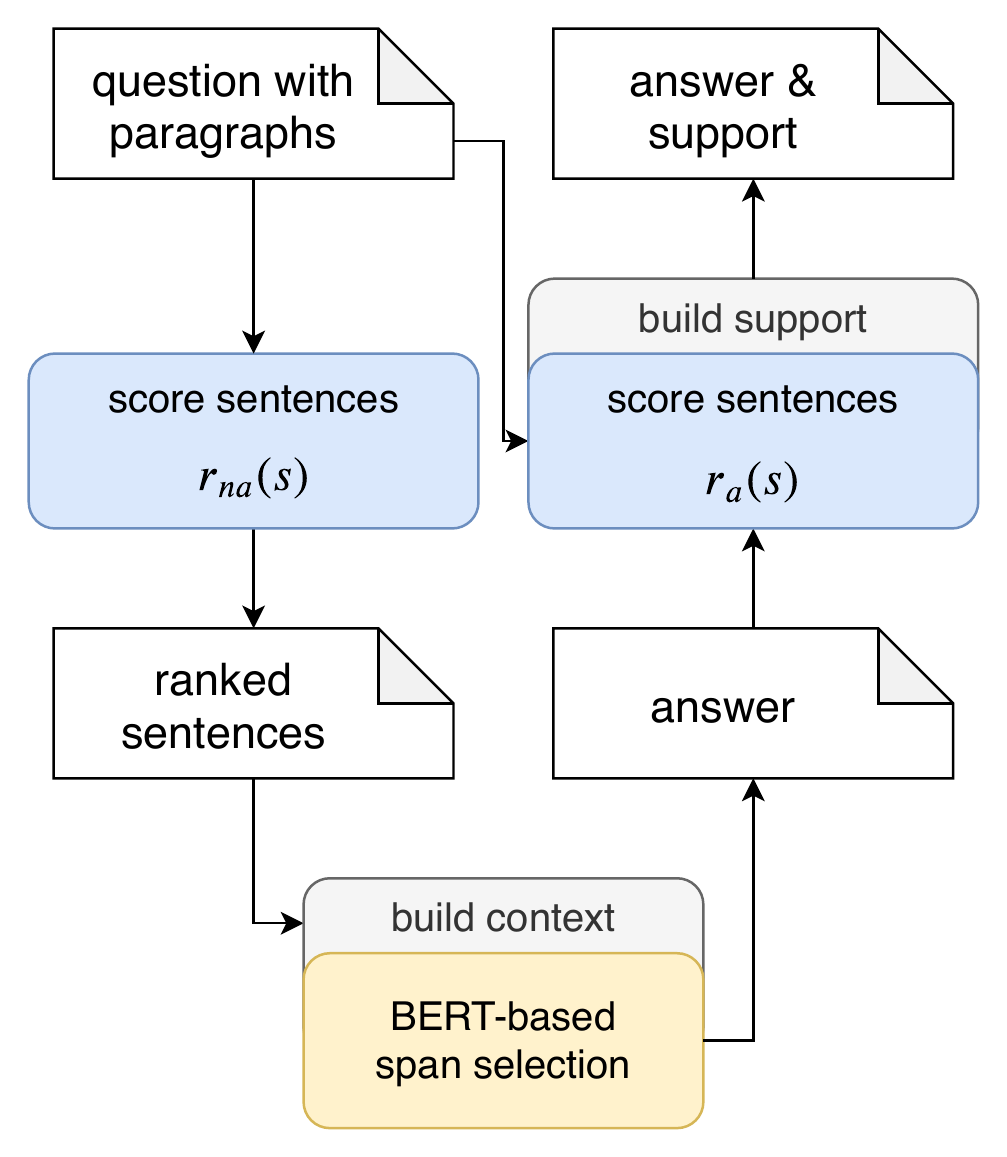}
    \caption{Overview of the \model model, with a question and context paragraphs as input. In both blue boxes, sentences are scored independently from one another. $r_{na}(s)$ and $r_a(s)$ use the same model architecture with different weights.}
    \label{fig:model}
\end{figure}

Due to the multi-hop nature of this dataset, it is natural to assume that the relevance of a sentence for a question would depend on the other sentences considered to be relevant. E.g., the relevance of ``Obama was born in Hawaii.''\ to the question ``Where was the 44$^\textrm{th}$ President of USA born?''\ depends on the other relevant sentence: ``Obama was the 44$^\textrm{th}$ President of US.'' As a result, many approaches designed for this task focus on \emph{jointly} identifying the relevant sentences (or paragraphs) via mechanisms such as cross-document attention, graph networks, and entity linking.

Our results question this basic assumption. We show that a simple model, \model (see Fig.~\ref{fig:model}), that first identifies relevant sentences from each paragraph \emph{independent} of other paragraphs, is surprisingly powerful on this task: in 90\% of the questions, \model's relevance module recovers all gold supporting sentences within the top-5 sentences. For QA, it uses a standard BERT~\cite{devlin2018bert} span prediction model (similar to current published models) on the output of this module. Additionally, \model exploits the inherent similarity between the relevant sentence identification task and the task of generating an explanation given an answer produced by the QA module: it uses the same architecture for both tasks. 

We show that this independent sentence scoring model results in a simple QA pipeline that outperforms all other BERT models in both `distractor' and `fullwiki' settings of \hpqa. In the distractor setting (10 paragraphs, including two gold, provided as context), \model achieves joint scores (answer and support prediction) within 0.75\% of the current state of the art. Even in the fullwiki setting (all 5M Wikipedia paragraphs as context), by combining our sentence selection approach with a commonly used paragraph selection approach~\cite{semanticmrs}, we outperform all previously published BERT models. In both settings, the only models scoring higher use \roberta~\cite{roberta}, a more robustly trained language model that is known to outperform BERT across various tasks. 

While our design uses multiple transformer models (now considered a standard starting point in NLP), our contribution is a simple pipeline without any bells and whistles, such as NER, graph networks, entity linking, etc.

The closest effort to \model is by \namecite{Min2019CompositionalQD}, who also propose a simple QA model for \hpqa. Their approach selects answers independently from each paragraph to achieve competitive performance on the question-answering subtask of \hpqa (they do not address the support identification subtask). We show that while relevant sentences can be selected independently, operating jointly over these sentences chosen from multiple paragraphs can lead to state-of-the-art question-answering results, outperforming independent answer selection by several points.

Finally, our ablation study demonstrates that the sentence selection module benefits substantially from using context from the corresponding paragraph. It also shows that running this module a second time, with the chosen answer as input, results in more accurate support identification.

\section{Related Work}

Most approaches for \hpqa attempt to capture the interactions between the paragraphs by either relying on cross-attention between documents or sequentially selecting paragraphs based on the previously selected paragraphs.

While \namecite{qfe} also use a standard Reading Comprehension (RC) model, they combine it with a special Query Focused Extractor (\textbf{QFE}) module to select relevant sentences for QA and explanation. The QFE module sequentially identifies relevant sentences by updating a RNN state representation in each step, allowing the model to capture the dependency between sentences across time-steps. \namecite{dfgn} propose a Dynamically Fused Graph Networks (\textbf{DFGN}) model that first extracts entities from paragraphs to create an entity graph, dynamically extract sub-graphs and fuse them with the paragraph representation. 
The Select, Answer, Explain (\textbf{SAE}) model~\cite{sae} is similar to our approach in that it also first selects relevant documents and uses them to produce answers and explanations. However, it relies on a self-attention over \emph{all} document representations to capture potential interactions. Additionally, they rely on a Graph Neural Network (GNN) to answer the questions. Hierarchical Graph Network (\textbf{HGN}) model~\cite{hgn} builds a hierarchical graph with three levels: entities, sentences and paragraphs to allow for joint reasoning.
\textbf{DecompRC}~\cite{decomprc} takes a completely different approach of learning to decompose the question (using additional annotations) and then answer the decomposed questions using a standard single-hop RC system.

Others such as \namecite{Min2019CompositionalQD} have also noticed that many \hpqa questions can be answered just based on a single paragraph. Our findings are both qualitatively and quantitatively different. They did not consider the support identification task, and showed strong (but not quite SoTA) QA performance by running a QA model independently on each paragraph. We, on the other hand, show that interaction is not essential for selecting relevant sentences but actually valuable for QA! Specifically, by using a context of relevant sentences spread across multiple paragraphs in steps 2 and 3, our simple BERT model outperforms previous models with complex entity- and graph-based interactions on top of BERT. We thus view \model as a different, stronger baseline for multi-hop QA.

In the fullwiki setting, each question has no associated context and models are expected to select paragraphs from Wikipedia. To be able to scale to such a large corpus, the proposed systems often select the paragraphs independent of each other. A recent retrieval method  in this setting is Semantic Retrieval~\cite{semanticmrs} where first the paragraphs are selected based on the question, followed by individual sentences from these paragraphs. However, unlike our approach, they do not use the paragraph context to select the sentences, missing key context needed to identify relevance.

\section{Pipeline Model: \model}

Our model works in three steps. First, we score individual sentences from an input set of paragraphs $D$ based on their relevance to the question. Second, we feed the highest-scoring sentences to a span prediction model to produce an answer to the question. Third, we score sentences from $D$ a second time to identify the supporting sentences using the answer.
 These three steps are implemented using the two modules described next in Sections~\ref{subsec:sentence-scoring-module} and~\ref{subsec:qa-module}.

\subsection{Sentence Scoring Module}
\label{subsec:sentence-scoring-module}

In the distractor setting, \hpqa provides 10 context paragraphs that have an average length of 41.4 sentences and 1106 tokens. This is too long for standard language-model based span-prediction---most models scale quadratically with the number of tokens, and some are limited to 512 tokens. This motivates selecting a few relevant sentences $E$ to reduce the size of the input to the span-prediction model without losing important context. In a similar vein, the support identification subtask of \hpqa also involves selecting a few sentences that best explain the chosen answer. We solve both of these problems with the same transformer-based sentence scoring module, with slight variation in its input.

Our sentence scorer uses the \ourlm model~\cite{devlin2018bert} trained with whole-word masking, with an additional linear layer over the \texttt{[CLS]} token. Here, whole word masking refers to a BERT variant that masks entire words instead of word pieces during pre-training. 

We score every sentence $s$ from every paragraph $p \in D$ independently by feeding the following sequence to the model: \texttt{[CLS] question [SEP] p [SEP] answer [SEP]}. This sequence is the same for every sentence in the paragraph, but the sentence being classified is indicated using a segment IDs: It is set to $1$ for tokens from the sentence and to $0$ for the rest. If a paragraph has more than 512 tokens, we restrict the input to the first 512. Each annotated support sentence forms a positive example and all other sentences from $D$ form the negative examples. Note that our classifier scores each sentence independently and never sees sentences from two paragraphs at the same time. (See Appendix \ref{appendix:training_sentence} for further detail.)

We train two variants of this model: (1) $r_{na}(s)$ is trained to score sentences given a question but no answer (\texttt{answer} is replaced with a \texttt{[MASK]} token); and (2) $r_a(s)$ is trained to score sentences given a question and its gold answer. We use $r_{na}(s)$ for relevant sentence selection and $r_a(s)$ for support identification (Sec.~\ref{subsec:distractor-setting}).

\subsection{Question Answering Module}
\label{subsec:qa-module}

To find answers to questions, we use \namecite{huggingface}'s implementation of \namecite{devlin2018bert}'s span prediction model. To achieve our best score, we use their \ourlm model with whole-word masking and  SQuAD~\cite{squad} fine-tuning.\footnote{While we use the model fine-tuned on SQuAD, ablations show that this only adds $0.2\%$ to the final score.}
We fine-tune this model on the \hpqa dataset with input QA context $E$ from $r_{na}(s)$. Since BERT models have a hard limit of 512 word-pieces, we use $r_{na}(s)$ to select the most relevant sentences that can fit within this limit, as described next. (See Appendix \ref{appendix:training_qa} for training details.)

To accomplish this, we compute the score $r_{na}(s)$ for each sentence in the input $D$. Then we add sentences in decreasing order of their scores to the QA context $E$, until we have filled no more than 508 word-pieces (incl. question word-pieces). For every new paragraph considered, we also add its first sentence, and the title of the article (enclosed in \texttt{<t></t>}). This ensures that our span-prediction model has the right co-referential information from each paragraph. We arrange these paragraphs in the order of their highest-scoring sentence, so the most relevant sentences come earlier  -- a signal that could be exploited by our model. The final four tokens are a separator, plus the words \texttt{yes}, \texttt{no}, and \texttt{noans}. This allows the model to answer yes/no comparison questions, or give no answer at all.

\begin{table*}
    \centering
    \small
    \setlength{\tabcolsep}{10pt}
    \setlength{\doublerulesep}{\arrayrulewidth}
    \begin{tabular}{l|cc|cc|cc}
    \hline \hline
    \multicolumn{1}{c|}{\multirow{2}{*}{QA Model}} & \multicolumn{2}{c|}{\T Answer} & \multicolumn{2}{c|}{Support} & \multicolumn{2}{c}{Joint} \\
    & \T EM & F1 & EM & F1 & EM & F1 \\
    \hline
    \T Single-paragraph~\cite{Min2019CompositionalQD} & -- & 67.08 & -- & -- & -- & -- \\
    QFE~\cite{qfe} & 53.70 & 68.70 & 58.80 & 84.70 & 35.40 & 60.60 \\
    DFGN~\cite{dfgn} & 55.66 & 69.34 & 53.10 & 82.24 & 33.68 & 59.86 \\
    SAE~\cite{sae} & 61.32 & 74.81 & 58.06 & 85.27 & 39.89 & 66.45 \\
    HGN~\cite{hgn} & -- & 79.69 & -- & \textbf{87.38} & -- & 71.45 \\
    \selectmod \model (Ours) & \textbf{67.75} & \textbf{81.21} & \textbf{60.72} & 86.97 & \textbf{44.35} & \textbf{72.26} \\
    \hline
    \T SAE (\roberta)~\cite{sae} & 67.70 & 80.75 & \textbf{63.30} & 87.38 & \textbf{46.81} & 72.75 \\
    HGN (\roberta)~\cite{hgn} & -- & 81.00 & -- & \textbf{87.93} & -- & \textbf{73.01} \\
    \hline \hline
    \end{tabular}
    \caption{\hpqa's distractor setting, Dev set. The bottom two models use larger language models than \model. 
    }
    \label{tab:distractor_results}
\end{table*}

\begin{table*}
    \centering
    \small
    \setlength{\tabcolsep}{10pt}
    \setlength{\doublerulesep}{\arrayrulewidth}
    \begin{tabular}{l|cc|cc|cc}
    \hline \hline
    \multicolumn{1}{c|}{\multirow{2}{*}{QA Model}} & \multicolumn{2}{c|}{\T Answer} & \multicolumn{2}{c|}{Support} & \multicolumn{2}{c}{Joint} \\
    & \T EM & F1 & EM & F1 & EM & F1 \\
    \hline
    \T QFE~\cite{qfe} & 28.66 & 38.06 & 14.20 & 44.35 & 8.69 & 23.10 \\
    SR-MRS~\cite{semanticmrs} & 45.32 & 57.34 & 38.67 & 70.83 & 25.14 & 47.60 \\
    \selectmod \model + SR-MRS (Ours) & \textbf{55.50} & \textbf{67.51} & \textbf{45.64} & \textbf{72.95} & \textbf{32.89} & \textbf{56.23} \\
    \hline
    \T
    HGN (\roberta) + SR-MRS ~\cite{hgn} & \textbf{56.71} & \textbf{69.16} & \textbf{49.97} & \textbf{76.39} & \textbf{35.36} & \textbf{59.86} \\
    \hline \hline
    \end{tabular}
    \caption{\hpqa's fullwiki setting, Test set. The bottom-most model uses a larger language model than \model.}
    \label{tab:fullwiki_results}
\end{table*}

\subsection{Bringing it Together: Distractor Setting}
\label{subsec:distractor-setting}

Given a question along with 10 distractor paragraphs $D$, we use the $r_{na}(s)$ variant of our sentence scoring module to score each sentence $s$ in $D$, again without looking at other paragraphs. In the second step, the selected sentences are fed as context $E$ into the QA module (as described in Section~\ref{subsec:qa-module}) to choose an answer. In the final step, to find sentences supporting the chosen answer, we use $r_a(s)$ to score each sentence in $D$, this time with the chosen answer as part of the input.\footnote{We simply append the answer string to the question even if it is ``yes'' or ``no''.}

We define the score $n(S)$ of a set of sentences $S \subset D$ to be the sum of the individual sentence scores; that is, $n(S) = \sum_{s \in S} r_a(s)$.\footnote{Note that $r_a(s)$ is the logit score and can be negative, so adding a sentence may not always improve this score.} In \hpqa, supporting sentences always come from exactly two paragraphs. We compute this score for all possible $S$ satisfying this constraint and take the highest scoring set of sentences as our support.

\subsection{Bringing it Together: Fullwiki Setting}
\label{subsec:fullwiki-setting}

Since there are too many paragraphs in the fullwiki setting, we use paragraphs from the SR-MRS system~\cite{semanticmrs} as our context $D$ for each question. On the Dev set, we found \model to perform best with a paragraph score threshold of $-8.0$ in MRS. Neither the sentence scorers $r_{na}(s), r_a(s)$ nor the QA module were retrained in this setting.

\section{Experiments}

We evaluate on both the distractor and fullwiki settings of \hpqa with the following goal: \emph{Can a simple pipeline model outperform previous, more complex, approaches?} We present the EM (Exact Match) and F1 scores on the evaluation metrics proposed for \hpqa: (1) answer selection, (2) support selection, and (3) joint score. 

Table~\ref{tab:distractor_results} shows that on the distractor setting, \model outperforms all previous models based on BERT, including HGN, which like us also uses whole word masking for contextual embeddings. Moreover, we are within 1 point of models that use RoBERTa embeddings---a much stronger language model that has shown improvements of 1.5 to 6 points in previous \hpqa models.

\model also performs better than the recent single-paragraph approach for the QA subtask \cite{Min2019CompositionalQD} by 14 points F1. While most of this gain comes from using a larger language model, \model scores 2 points higher even with a language model of the same size (BERT-Base).

We observe a similar trend in the fullwiki setting (Table~\ref{tab:fullwiki_results}) where \model again outperforms previous approaches (except HGN with RoBERTa). While we rely on retrieval from SR-MRS~\cite{semanticmrs} for our initial paragraphs, we outperform the original work.
We attribute this improvement to two factors: our sentence selection capitalizing on the sentence's paragraph context leading to better support selection, and a better span selection model leading to improved QA.

\subsection{Ablation}

\begin{table}[t]
    \centering
    \small
    \setlength{\doublerulesep}{\arrayrulewidth}
    \begin{tabular}{l|c|c|c}
    \hline \hline
    \T                            & top-\emph{n} & Sup F1 & Ans F1 \\
    \hline
    \T B-Base w/o context         & 10 &  74.45 &  78.59 \\
    B-Base w/ context             &  6 &  83.15 &  80.92 \\
    + B-Large ($r_{na}(s)$)       & \textbf{5} & 85.35 & \textbf{81.21} \\
    \hline
    \T w/ answers ($r_a(s)$)      &  5 & \textbf{86.97} & -- \\
    Oracle                        &  3 & -- & -- \\
    \hline \hline
    \end{tabular}
    \caption{Ablation study on sentence selection in the distractor setting. top-\emph{n} indicates the number of sentences required to cover the annotated support sentences in 90\% of the questions.}
    \label{tab:sentence_selection}
\end{table}

To evaluate the impact of context on our sentence selection model in isolation, we look at the number of sentences that score at least as high as the lowest-scoring annotated support sentence. In other words, this is the number of sentences we must send to the QA model to ensure all annotated support is included. Table~\ref{tab:sentence_selection} shows that providing the model with the context from the paragraph gives a substantial boost on this metric, bringing it down from 10 to only 6 when using BERT-Base (an oracle would need 3 sentences). It further shows that this boost carries over to the downstream tasks of span selection and choosing support sentences (improving it by 9 points to 83\%). Finally, the table shows the value of running the sentence selection model a second time: with BERT-Large, $r_a(s)$ outperforms $r_{na}(s)$ by 1.62\% on the Support F1 metric.

Looking deeper, we analyzed the accuracy of our third stage, $r_a(s)$, as a function of the correctness of the QA stage. When QA finds the correct gold answer, $r_a(s)$ obtains the right support in 65.9\% of the cases. If the answer from QA is incorrect, the success rate of $r_a(s)$ is only 50.9\%.

\section{Conclusion}

Our work shows that on the \hpqa tasks, a simple pipeline model can do as well as or better than more complex solutions. Powerful pre-trained models allow us to score sentences one at a time, without looking at other paragraphs. By operating jointly over these sentences chosen from multiple paragraphs, we arrive at answers and supporting sentences on par with state-of-the-art approaches. This result shows that retrieval in \hpqa is not itself a multi-hop problem, and suggests focusing on other multi-hop datasets to demonstrate the value of more complex techniques.

\bibliographystyle{acl_natbib}
\bibliography{references}

\newpage
\appendix

\section{Appendix}

\subsection{Training the sentence scoring model}
\label{appendix:training_sentence}
Both $r_{na}(s)$ and $r_a(s)$ are trained the same way. We use the 90447 questions from the \hpqa training set, shuffle them, and train for 4 epochs. Both models are trained in the distractor setting only, but evaluated in both settings. We construct positive and negative examples by choosing the two paragraphs containing the annotated support sentences, plus two more randomly chosen paragraphs. All sentences from the chosen paragraphs become instances for the model.

During training, we follow the fine-tuning advice from~\cite{devlin2018bert}, with two exceptions. We ramp up the learning rate from $0$ to $10^{-5}$ over the first 10\% of the batches, and then linearly decrease it again to $0$.

To avoid biasing the training towards questions with many context sentences, we create batches at the question level. Three questions make up one batch, regardless of how many sentences they contain. We cap the batch size at 5625 tokens for practical purposes. If a batch exceeds this size, we drop sentences at random until the batch is small enough.
As is standard for BERT classifiers, we use a cross-entropy loss with two classes, one for positive examples, and one for negative examples.

\subsection{Training the span prediction model}
\label{appendix:training_qa}
We train the BERT span prediction model on the output paragraphs from $r_{na}(s)$. We use a batch size of 16 questions and maximum sequence length of 512 word-pieces. We use the same optimizer settings as the sentence selection model with an additional weight decay of $0.01$. The model is trained for a fixed number of epochs (set to 3) and the final model is used for evaluation.
Under the hood, this model consists of two classifiers that run at the same time. One finds the first token of potential spans, and one finds the last token of potential spans. Each classifier uses a cross entropy loss. The final loss is the average loss of the two classifiers.
We train one model on the output from our best $r_{na}(s)$ selection model and use it in all our experiments (and ablations).

\end{document}